%% file: main.tex
\documentclass[10pt,twocolumn,letterpaper]{article}

\usepackage[pagenumbers]{cvpr} 


\definecolor{cvprblue}{rgb}{0.21,0.49,0.74}
\usepackage[pagebackref,breaklinks,colorlinks,allcolors=cvprblue]{hyperref}

\def\paperID{9999999} 
\def\confName{CVPR}
\def\confYear{2025}

\title{Dual Degradation-Inspired Deep Unfolding Network for Low-Light Image Enhancement}

\author{
\textbf{Huake Wang$^{1}$,
Xingsong Hou$^1$\thanks{Corresponding author.},
Chengxu Liu$^{1}$,
Kaibing Zhang$^2$, 
Xiangyong Cao$^1$,
Xueming Qian$^1$}
\smallskip 
\\
$^1$Xi’an Jiaotong University, $^2$Xi'an Polytechnic University
\smallskip 
\\
\tt\small\{wanghuake, chengxuliu\}@stu.xjtu.edu.cn
\tt\small\{houxs, caoxiangyong, qianxm\}@mail.xjtu.edu.cn
}

\begin{document}
\maketitle
\input{sec/0_abstract}    
\input{sec/1_intro}

\input{sec/2_relatedwork}

\input{sec/3_method}

\input{sec/4_results}
\input{sec/5_conclusion}
{
    \small
    \bibliographystyle{ieeenat_fullname}
    \bibliography{main}
}

\input{sec/X_suppl}

\end{document}

%% file: sec/0_abstract.tex
\begin{abstract}
Although low-light image enhancement has achieved great stride based on deep enhancement models, most of them mainly stress on enhancement performance via an elaborated black-box network and rarely explore the physical significance of enhancement models. Towards this issue, we propose a \textbf{D}ual degr\textbf{A}dation-in\textbf{S}pired deep \textbf{U}nfolding network, termed DASUNet, for low-light image enhancement. Specifically, we construct a dual degradation model (DDM) to explicitly simulate the deterioration mechanism of low-light images. It learns two distinct image priors via considering degradation specificity between luminance and chrominance spaces. To make the proposed scheme tractable, we design an alternating optimization solution to solve the proposed DDM. Further, the designed solution is unfolded into a specified deep network, imitating the iteration updating rules, to form DASUNet. Based on different specificity in two spaces, we design two customized Transformer block to model different priors. Additionally, a space aggregation module (SAM) is presented to boost the interaction of two degradation models. Extensive experiments on multiple popular low-light image datasets validate the effectiveness of DASUNet compared to canonical state-of-the-art low-light image enhancement methods. Our source code and pretrained model will be publicly available.
\end{abstract}

%% file: sec/1_intro.tex
\section{Introduction}
\label{sec:intro}

Due to lacking of suitable light source, images shot at the surrounding exhibit poor visibility, weak contrast, and unpleasant noise. It not only prejudices the visual comfort of the observer, but also impairs the analysis performance of down-stream high-level vision tasks, e.g., detection~\cite{1, 2, 3,75}, tracking~\cite{4, 5}, recognition~\cite{6, 7}, and segmentation~\cite{8, 9, 10}. Hence, how to restore low-light images is an urgent problem to be solved.

\begin{figure}
  \centering
  \centerline{\includegraphics[width=0.48\textwidth]{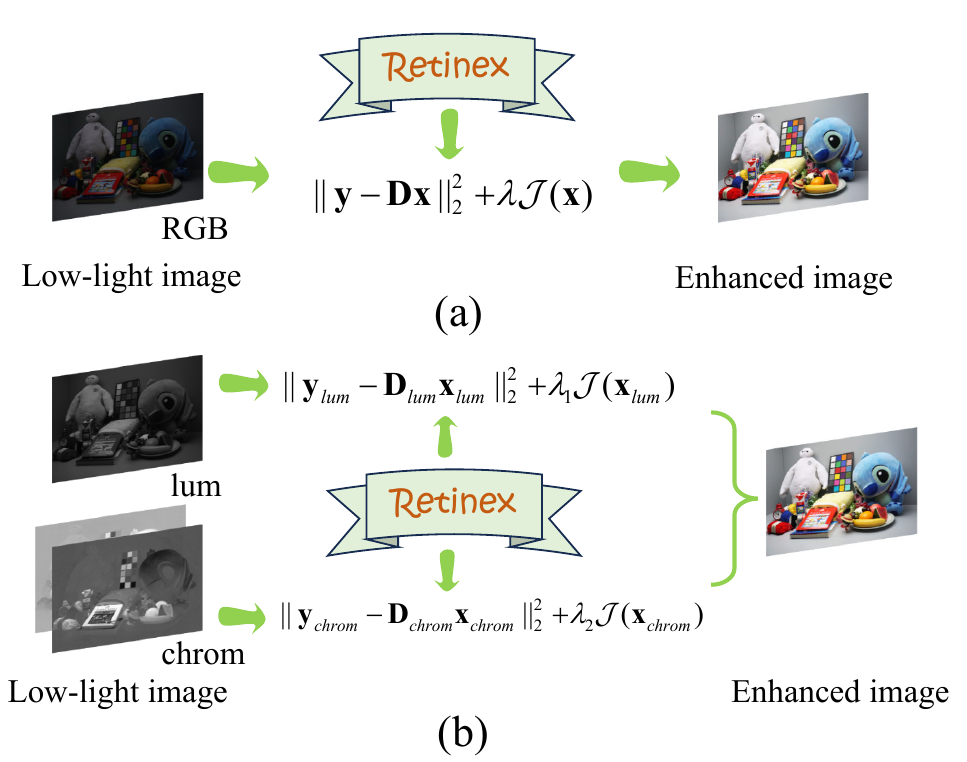}}
  \caption{The comparison between existing unfolding methods and our proposed method. (a) the existing unfolding methods; (b) our proposed method.}
\vspace*{-5mm}
\end{figure}

Low-light image enhancement refers to improving the brilliance of image content and rendering the radiance of image focus. Generally, an image is considered as the product of an luminance layer and a reflectance layer based on Retinex theory~\cite{11}. Mathematically, it can be defined as $\mathbf{y}=\mathbf{D} \otimes \mathbf{x}$, where $\otimes$ denotes the element-level multiplication, $\mathbf{D}$ and $\mathbf{x}$ are the illumination layer and the reflectance layer which imply the extrinsic brightness and the intrinsic property of an object, respectively. According to the statement, some traditional methods attempt to separate reflectance maps from low-light images as enhanced results via Gaussian filter~\cite{11}, multi-scale Gaussian filter~\cite{12}, or luminance structure prior~\cite{13}. However, they could produce some unnatural appearances. Alternatively, some researchers simultaneously increase luminance component and restore reflectance component to further adjust under-exposure images via bright-pass filter~\cite{14} and weighted variational model~\cite{15}. Unfortunately, this way could under-enhance or over-enhance these images with uneven brightness. Besides, several additional penalty constraints, e.g., noise prior~\cite{16}, low-rank constraint~\cite{17}, structure prior~\cite{18, 19}, are combined into the final optimization objective function to further improve the image objective quality. However, hand-crafted priors could not well encapsulate the various real low-light scenes. In short, above traditional methods are hard to yield impressive enhancement performance.

Fueled by convolutional neural network, a large group of researchers have attempted to train high-efficient low-light image enhancement models from a sea of data. Deep black-box methods directly learn a powerful mapping relationship from low-light images to normal-light images via denoising autoencoder~\cite{20}, multi-scale framework~\cite{21,22,51,70}, normalizing flow~\cite{23}, Transformer~\cite{24, 25}, deep unsupervised framework~\cite{26, 27}, and so on. However, they focus more on enhancement performance at the expense of the interpretability and physical degradation principle, which could heavily hinder the performance improvements. In the light of imaging theory, Retinex-based deep models are proposed to decompose natural images to luminance layers and reflectance layers, which could be trained via reflectance layer consistency constraint~\cite{28, 29}, illumination smoothness~\cite{30}, and semantic-aware prior~\cite{31}. However, the above methods only mimic ill-posed decomposition in form but lack accurate luminance references. Moreover, Deep unfolding methods~\cite{32, 33, 45} unrolled Retinex theory into a convolutional network to combine the physical principle and powerful representation ability of deep model. As shown in Fig. 1, the above unfolding methods directly conduct on RGB space. Deep unfolding methods operating on other color space with specific advantages, such as, YCbCr~\cite{72,34,35}, remian unexplored.


\begin{figure*}
  \centering
  \centerline{\includegraphics[width=0.88\textwidth]{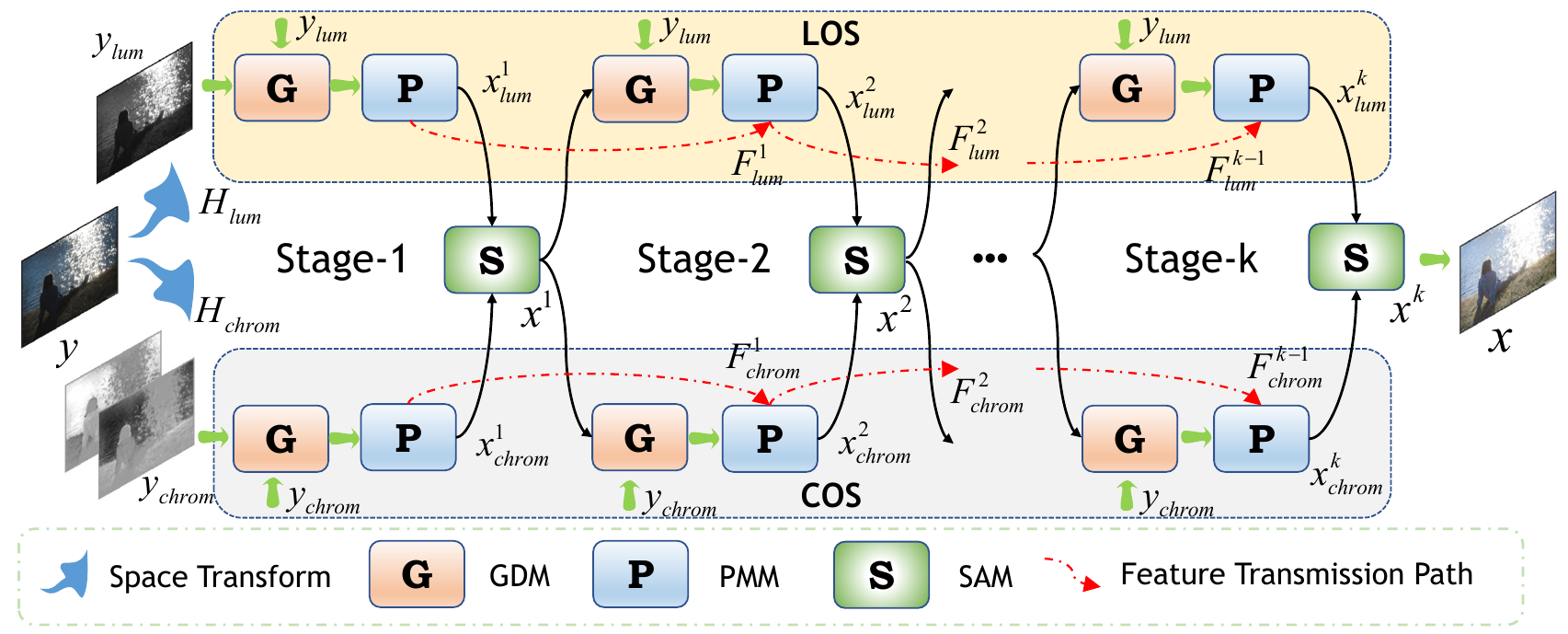}}
  \caption{The architecture of our DASUNet. The low-light image $\mathbf{y}$ is transformed to luminance space $\mathbf{y}_{lum}$ and chrominance space $\mathbf{y}_{chrom}$. Then, they are respectively fed into luminance optimization stream (LOS) and chrominance optimization stream (COS). After $k$ stages, the enhanced image $\mathbf{x}$ is produced.}
\vspace*{-5mm}
\end{figure*}

To push the frontiers of deep unfolding-based image enhancement, we propose a \textbf{D}ual degr\textbf{A}dation-in\textbf{S}pired deep \textbf{U}nfolding network, termed \textbf{DASUNet}, for low-light image enhancement, which is shown in Fig. 2. The motivation originates from the degradation specificity of low-light images between luminance and chrominance spaces~\cite{72,34,35, 85}. On this basis, we formulate the task of low-light image enhancement as the optimization of dual degradation model, which inherits the physical deterioration principle and interpretability. Further, an alternating optimization solution is designed to solve the proposed dual degradation model. Then, the iterative optimization rule is unfolded into a deep model, composing DASUNet, which enjoys the strengths of both the physical model and the deep network. Based on the differences of luminance and chrominance spaces, we customized two different prior modeling modules (PMM) to learn different prior information. In the luminance channel, we design a luminance adjustment Transformer to modulate brightness strength. While in the chrominance channel, a Wavelet decomposition Transformer is proposed to model high-frequency and low-frequency information leveraging the advantages of convolutions and Transformer~\cite{36,37,38, 39}.
Besides, we design a space aggregation module (SAM) to yield the clear images, which combines the reciprocity of dual degradation priors. We perform extensive experiments on several datasets to analyze and validate the effectiveness and superiority of our proposed DASUNet compared to other state-of-the-art methods.

In a nutshell, the main contributions of this work are summarized as follows:

\begin{itemize}
\item We propose a dual degradation model based on degradation specificity of low-light images on different spaces. It is unfolded to form dual degradation-inspired deep unfolding network for low-light image enhancement, which can jointly learn two degradation priors from luminance space and chrominance space. More importantly, dual degradation model empowers DASUNet with explicit physical insight, which improves the interpretability of enhancement model.
\item We design two different prior modeling modules, luminance adjustment Transformer and Wavelet decomposition Transformer, to obtain degradation-aware priors based on degradation specificity of different spaces.
\end{itemize}

%% file: sec/2_relatedwork.tex
\section{Related Work}

\subsection{Low-light image enhancement}

Recently, deep low-light image models have highly outperformed the traditional enhancement models not only in performance but also in running time. As a pioneering work, LLNet~\cite{20} lightened low-light images using the denoising auto-encoder. After that, a large number of researchers~\cite{21,22,24,25,40,41,26,42,43} built the mapping relation between low-light images and normal-light images via different blocks or manners. 
Alternatively, some methods~\cite{28,29,30} attempted to decompose low-light images into reflectance and luminance maps via a data-driven deep model built on Retinex theory. Besides, more researchers~\cite{26,27,32,45} have started to construct an unsupervised enhancement model to alleviate the shortage of paired images. However, above methods generally connot consider the physical imaging principle. 


\subsection{Deep unfolding method}
Deep unfolding model inherits the advantages of the interpretability of physical models and the powerful representation capability of deep models driven by a large number of images, which has shown prominent performance in many vision tasks, e.g., super resolution~\cite{48,47,46,73} and compressive sensing~\cite{50,49,51}. Concretely speaking, deep unfolding model could optimize the iterative solver of physical models via some convolution networks in an end-to-end manner. 
In the field of low-light image enhancement, URetinex~\cite{33} proposed a Retinex theory-inspired unfolding model, however too many constraint terms hindered its performance. RUAS~\cite{32} and SCI~\cite{45} adopted a reference-free training mechanism to supervise the unfolding framework, limiting their performance. Unlike them that only learn a universal prior information, our method unfolds a novel dual degradation model based on space difference into an elaborated network, which learn two complementary and effective prior information to restore low-light images.



%% file: sec/3_method.tex
\section{Methodology}
In this section, we firstly introduce our designed dual degradation model (DDM) as the objective function. Then, an iterative optimization solution is designed as the solver of DDM. Moreover, we elaborate on dual degradation-inspired unfolding network (DASUNet) for low-light image enhancement, which is delineated on Fig. 2. 

\subsection{Dual Degradation Model}
Fundamentally, given a low-light image $\mathbf{y}\in \mathbb{R}^{H\times W\times 3} $, this paper aim to recover a normal-light image $\mathbf{x}\in \mathbb{R}^{H\times W\times 3} $ from it. $H\times W\times 3$ denotes the spatial dimension of an image. According to Retinex theory, the degradation model of low-light images can be described as:
\begin{equation}
\mathbf{y}=\mathbf{D}\otimes\mathbf{x}, 
\end{equation}
where $\mathbf{D}$ is the illumination layer, which in fact is a hybrid degradation operator that may include illuminance deterioration, color distortion, and detail loss ~\cite{30,55}. Hence, we can recover $\mathbf{x}$ by minimizing the following energy function:
\begin{equation}
\min_{\mathbf{x}} \frac{1}{2} || \mathbf{y}-\mathbf{Dx}|| ^{2}_{2} + \lambda \mathcal{J}(\mathbf{x}), 
\end{equation}
where $|| \mathbf{y}-\mathbf{Dx}|| ^{2}_{2}$ is the data fidelity term, $\mathcal{J}(\mathbf{x})$ represents the degradation prior term, and $\lambda$ denotes the hyperparameter weighting the significance of prior term. Due to the ill-posedness of degradation model, many hand-crafted prior items, e.g., non-local similarity ~\cite{56,57} and low-rank prior ~\cite{58}, are introduced to approximate a desired solution. However, above priors are hard to depict the universal structure of natural images. Recently, deep denoising prior ~\cite{59,60} is proposed to characterize the image-generic skeleton in a data-driven manner, which show more competitive performance in many low-level vision tasks. Hence, this paper also adopts deep prior as the regularization constraint.

Traditional degradation models ~\cite{58,56} generally conduct in a single image space, e.g., RGB or Y, which show similar deterioration type. However, low-light images show diverse deterioration types, leading to less effectiveness of above degradation model for their enhancement. Of note, we find the degradation specificity between luminance and chrominance spaces from low-light images ~\cite{34,35}, which is demonstrated in Fig. 3. One can see obvious degradation difference in luminance and chrominance spaces, which inspires us to design DDM to describe different deterioration process. DDM can be obtained via transforming Eq. (2):
\begin{equation}\small
\begin{aligned}
\min_{\mathbf{x}_{\text{lum}},\mathbf{x}_{\text{chrom}}} &\frac{1}{2} || \mathbf{y}_{lum}-\mathbf{D}_{lum}\mathbf{x}_{lum}|| ^{2}_{2} + \lambda _{1} \mathcal{J}(\mathbf{x}_{lum})  \\
&+ \frac{1}{2} || \mathbf{y}_{chrom}-\mathbf{D}_{chrom}\mathbf{x}_{chrom}|| ^{2}_{2}+ \lambda _{2} \mathcal{J}(\mathbf{x}_{chrom})  \\
\end{aligned}
\end{equation}
where $\mathbf{D}_{lum}$ and $\mathbf{D}_{chrom}$ refer to luminance and chrominance degradation operators, $\mathbf{x}_{lum} \in \mathbb{R}^{H\times W\times 1} $ and $\mathbf{y}_{lum} \in \mathbb{R}^{H\times W\times 1} $ denote luminance component of normal-light and low-light images, $\mathbf{x}_{chrom} \in \mathbb{R}^{H\times W\times 2} $ and $\mathbf{y}_{chrom} \in \mathbb{R}^{H\times W\times 2} $ denote chrominance component of normal-light and low-light images, $\mathcal{J}(\mathbf{x}_{lum})$ and $\mathcal{J}(\mathbf{x}_{chrom})$ denote the luminance space prior and chrominance space prior, $\lambda _{1}$ and $\lambda _{2}$ are two weight parameters. More specifically, $\mathbf{x}_{lum} $, $\mathbf{y}_{lum}$, $\mathbf{x}_{chrom}$, and $\mathbf{y}_{chrom}$ can be obtained via:
\begin{equation} \small
\begin{aligned}  
\mathbf{x}_{lum},\mathbf{y} _{lum} &=\mathbf{H}_{lum}(\mathbf{x},\mathbf{y}), \\
 \mathbf{x}_{chrom},\mathbf{y} _{chrom}&=\mathbf{H}_{chrom}(\mathbf{x},\mathbf{y}),
\end{aligned}
\end{equation}
where $\mathbf{H}_{lum}$ and $\mathbf{H}_{chrom}$ signify the luminance and chrominance transform. For compactness, we omit $\mathbf{x}_{lum},\mathbf{y} _{lum}=\mathbf{H}_{lum}(\mathbf{x},\mathbf{y})$ and $\mathbf{x}_{chrom},\mathbf{y} _{chrom}=\mathbf{H}_{chrom}(\mathbf{x},\mathbf{y})$ in the following.  Moreover, we can easily transform luminance and chrominance components to original RGB space via the inverse transform of $\mathbf{H}_{lum}$ and $\mathbf{H}_{chrom}$.  DDM disentangles the intricate low-light degradation process into two easy independent degradation operators, which significantly reduce the modelling difficulty of prior term.

\begin{figure}[!t]
  \centering
  \centerline{\includegraphics[width=0.45\textwidth]{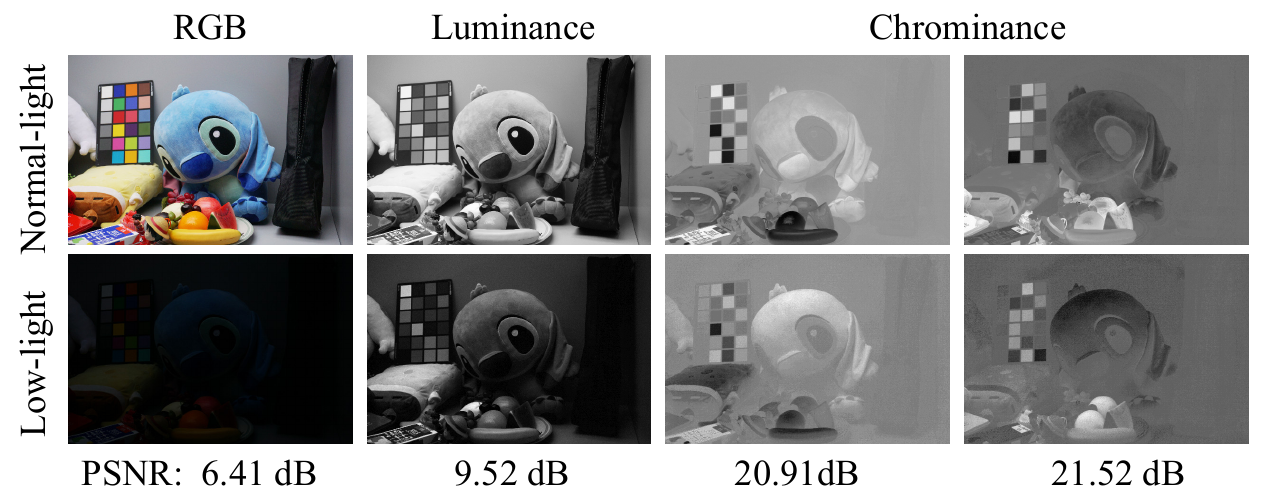}}
  \caption{The comparison of luminance and chrominance spaces. We can see that there are different distoration degrees in luminance and chrominance from quantitative and qualitative perspectives.}
\vspace*{-3mm}
\end{figure}

\subsection{Optimization Solution}
To achieve accurate normal-light images from low-light images, we design an alternate optimization solution to solve DDM. Firstly, we separate Eq. (3) into two independent objects to facilitate the optimization solution: 
\begin{equation} \small
\min_{\mathbf{x}_{\text{lum}}} \frac{1}{2} || \mathbf{y}_{lum}-\mathbf{D}_{lum}\mathbf{x}_{lum}|| ^{2}_{2} + \lambda _{1} \mathcal{J}(\mathbf{x}_{lum}), 
\end{equation}
\vspace*{-5mm}
\begin{equation} \small
\min_{\mathbf{x}_{\text{chrom}}} \frac{1}{2} || \mathbf{y}_{chrom}-\mathbf{D}_{chrom}\mathbf{x}_{chrom}|| ^{2}_{2}+ \lambda _{2} \mathcal{J}(\mathbf{x}_{chrom}), 
\end{equation}
where Eq. (5) represents luminance degradation model and Eq. (6) refers to chrominance degradation model. They can be parallelly updated in luminance and chrominance spaces.

Taking Eq. (5) as example, it can be solved via proximal gradient algorithm~\cite{49,50}, which has been demonstrated its effectiveness in many inverse problems. Specifically, we can alternatively deduce the following updating rule to obtain an approximate solution:
\begin{subequations} 
\begin{align}
&u^{k}=\mathbf{x}^{k-1}_{lum}-\rho ^{k}\mathbf{D}_{lum}^{T}(\mathbf{D}_{lum} \mathbf{x}_{lum}^{k-1}-\mathbf{y}_{lum}),\\
&\mathbf{x}^{k}_{lum}=Prox_{\lambda_{1}}(u^{k}), 
\end{align}
\end{subequations}
where $\mathbf{x}^{k}_{lum}$ is the $k$-th updating solution, $\rho ^{k}$ denotes the updating stepsize, $\mathbf{D}_{lum}^{T}$ is the transpose of $\mathbf{D}_{lum}$, $Prox_{\lambda_{1}}$ represents the proximal operator. Generally, the first item is called as the gradient descent and the second item is thought as the proximal mapping.

\begin{figure*} [!t]
  \centering
  \centerline{\includegraphics[width=0.9\textwidth]{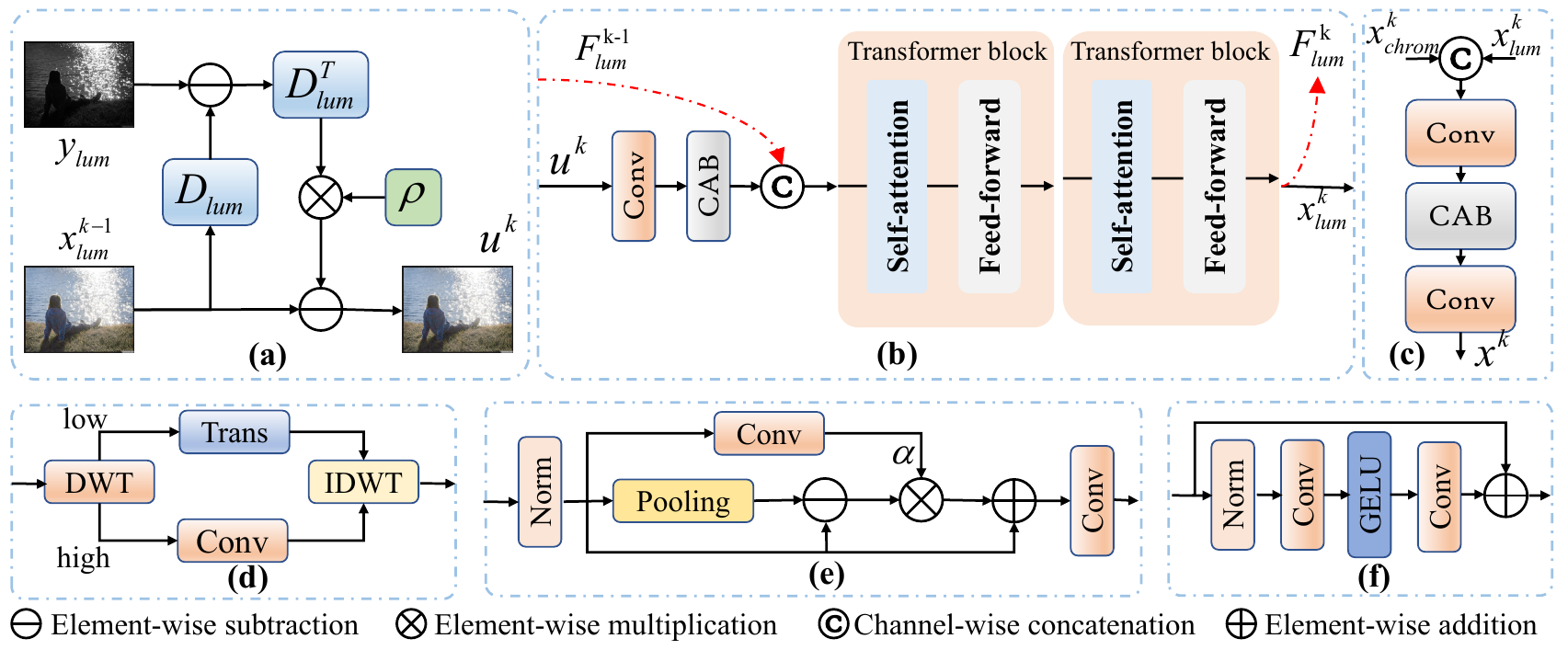}}
  \caption{The illustration of network structure of each phase. (a) Gradient Descent Module (GDM); (b) Prior Modelling Module (PMM); (c) Space Aggregation Module (SAM); (d) Wavelet decomposition Transformer; (e) luminance adjustment Transformer; (f)  Feed-forward layer.}
\vspace*{-5mm}
\end{figure*}

Similarly, chrominance degradation model is also solved by:%
\begin{subequations} \small
\begin{align}
&v^{k}=\mathbf{x}^{k-1}_{chrom}-\eta ^{k}\mathbf{D}_{chrom}^{T}(\mathbf{D}_{chrom} \mathbf{x}_{chrom}^{k-1}-\mathbf{y}_{chrom}),\\
&\mathbf{x}^{k}_{chrom}=Prox_{\lambda_{2}}(v^{k}).
\end{align}
\end{subequations}

Finally, we can obtain the output results during $k$-th updating process via:
\begin{equation} 
\mathbf{x}^{k} = [\mathbf{H}^{T}_{lum}(\mathbf{x}^{k}_{lum}); \mathbf{H}^{T}_{chrom}(\mathbf{x}^{k}_{chrom})], 
\end{equation}
where $\mathbf{H}^{T}_{lum}(\cdot)$ and $\mathbf{H}^{T}_{chrom}(\cdot)$ are the inverse transformation of luminance and chrominance transform. $[\cdot;\cdot]$ represents the space merging operation.

\subsection{Dual Degradation-Inspired Deep Unfolding Network}

Based on above designed iteration optimization solution, we unfold each iterative step into corresponding modules to construct our DASUNet, which is shown in Fig. 2. Specifically, our DASUNet is composed of $k$ stages, each of which contains two parts, luminance optimization stream (LOS) and chrominance optimization stream (COS). Furthermore, we design a space aggregation module (SAM) to combine the output of each stage of COS and LOS, which can interact the complementary features learned from luminance and chrominance spaces to produce more high-quality normal-light restored images.

\textbf{Luminance Optimization Stream.} LOS is composed of gradient descent modules (GDM) and prior modeling modules (PMM), which are unfolded from Eq. (7). Taking $k$-th stage for example, GDM adopts low-light luminance component $\mathbf{y}_{lum}$ and the $(k-1)$-th restored result $\mathbf{x}_{lum}^{k-1}$ as inputs, which is depicted in Fig. 4(a). A non-trivial problem is the construction of the degradation operator $\mathbf{D}_{lum}$ and its transpose $\mathbf{D}_{lum}^{T}$. Inspired by~\cite{49, 50}, we employ a residual convolution block to simulate $\mathbf{D}_{lum}$ and $\mathbf{D}_{lum}^{T}$. Their structure all adopt Conv-PReLU-Conv, and the channel numbers of two Convs are set to $n$ and 1. For $\rho^{k-1}$, it is a learnable parameter that is updated in each epoch and initialized to $0.5$. 

Another important module is PMM, which is also thought as a denoising network in many works~\cite{59,60}. Previous methods usually exploit convolution block, e.g., U-net and Resblock, to learn a universal image prior driven by tremendous specified-type images. However, they only extract local image prior via stacking more convolutions but cannot learn more representative long-range image structure. Recently, Transformer has been demonstrated its superiority for long-range information modeling in many vision tasks~\cite{36,37,38,39}. Motivated by them, we design a novel Transformer-based PMM to model long-range structural knowledge, as shown in Fig. 5(b). Moreover, some researchers~\cite{52,53} found the information flow among adjacent stages limits the performance of deep unfolding networks and propose a memory-augmented mechanism to facilitate the information flow across stages. Consequently, we build a high-way feature transmission path among adjacent stages without any convolution to aggravate the model efficiency, as shown in Fig. 2.

\begin{figure*} [!t]
  \centering
  \centerline{\includegraphics[width=0.98\textwidth]{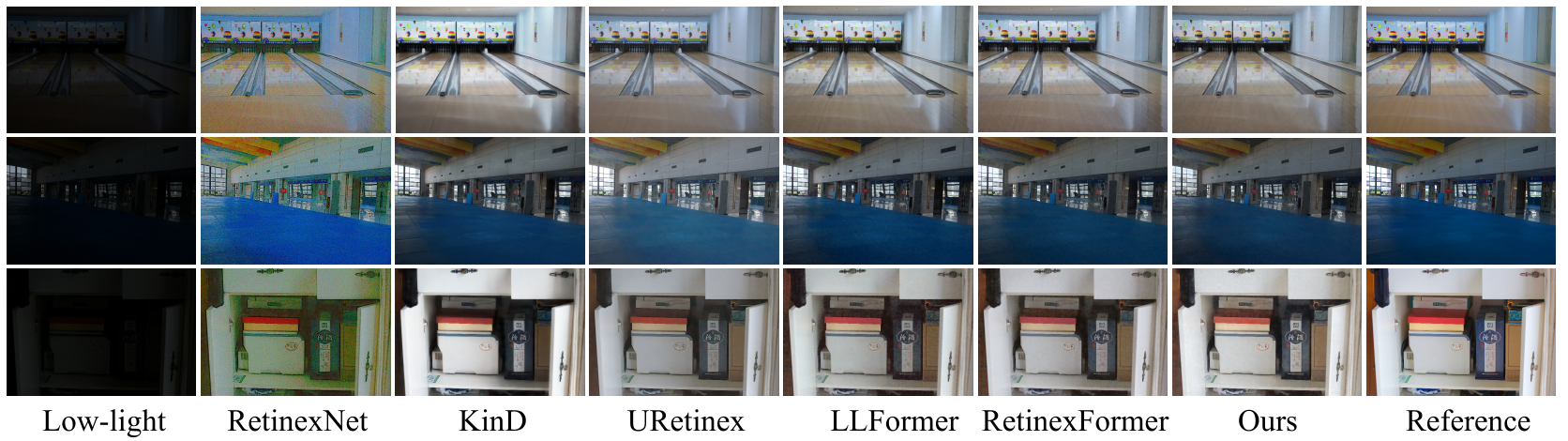}}
  \caption{The visual comparison on LOL dataset.}
  \label{fig:4}
\end{figure*}

\begin{figure*} [!t]
  \centering
  \centerline{\includegraphics[width=0.98\textwidth]{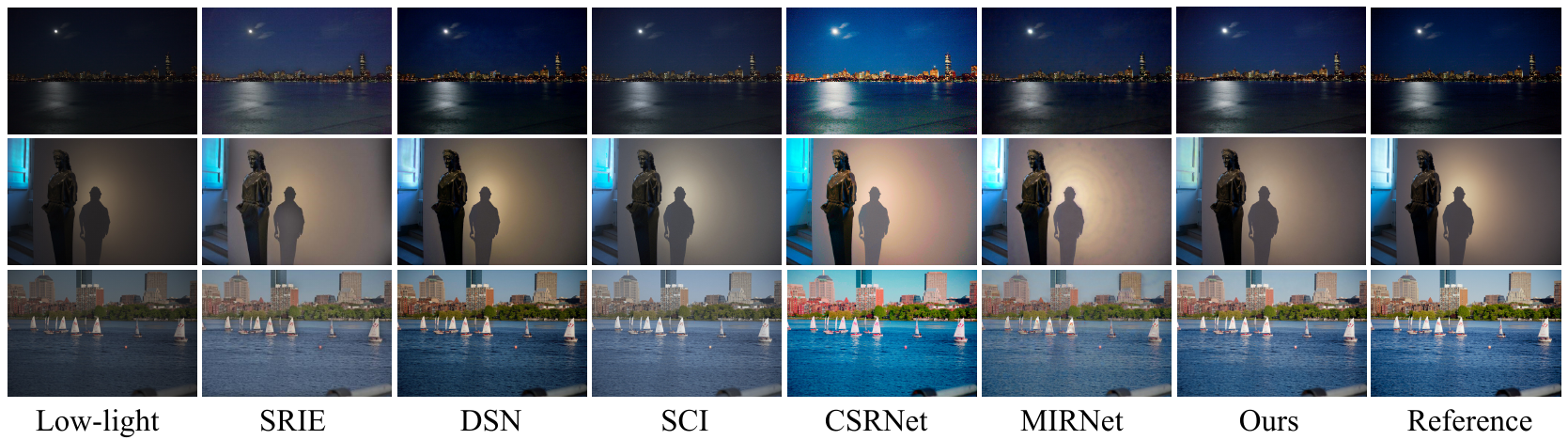}}
  \caption{The visual comparison on MIT-Adobe FiveK dataset.}
\end{figure*}

Specifically, the input is first fed into a convolution with kernel size of $3$ and a channel attention block (CAB)~\cite{61} to extract shallow features. Then, two $3\times3$ convolutions with $n$ channels are used to aggregate the shallow features and high-way features from previous stage. Next, two Transformer blocks are used to learn long-range prior features, each of which includes a self-attention layer and a feed-forward layer.
Finally, the learned features are sent into high-way features transmission path to flow into the next stage and SAM to produce the restored results of this stage. Overall, LOS can be described as:
\begin{subequations} 
\begin{align}
&u^{k}= \text{GDM}_{\lambda_{1}}(\mathbf{x}_{lum}, \mathbf{x}^{k-1}_{lum}),\\
&\mathbf{x}^{k}_{lum}, \mathbf{F}^{k}_{lum} = \text{PMM}_{\lambda_{1}}(u^{k}, \mathbf{F}^{k-1}_{lum}).
\end{align}
\end{subequations}

In PMM of LOS, we propose a novel luminance adjustment Transformer (LAT) to replace the Transformer block of PMM. LAT is shown in Fig. 4(e). Considering the local smoothness of brightness, we construct a local brightness difference via the pooling layer and pixel-wise subtraction. Then, we learn a brightness scale factor $\alpha$ exploiting a convolution layer to adjust the brightness. Finally, we use a convolution to obtain the final luminance prior information.

\textbf{Chrominance Optimization Stream.} COS has the same structure as LOS except for PMM. In COS, we customized a new self-attention layer, Wavelet decomposition Transformer, based on the characteristic of chrominance channels of low-light images, which is shown in Fig. 4(d). The input features are split into low-frequency part and high-frequency part via Wavelet filters. Low-frequency part is inputted into a Transformer block to extract long-range degradation knowledge. Since high-frequency part is sparse, it passed into a convolution block to learn local degradation prior. The design can facilitate the modeling of detail prior and reduce the computation overhead.

Similarly, COS is represented as:
\begin{subequations} 
\begin{align}
&v^{k}= \text{GDM}_{\lambda_{2}}(\mathbf{x}_{chrom}, \mathbf{x}^{k-1}_{chrom}),\\
&\mathbf{x}^{k}_{chrom}, \mathbf{F}^{k}_{chrom} = \text{PMM}_{\lambda_{2}}(v^{k}, \mathbf{F}^{k-1}_{chrom}).
\end{align}
\end{subequations}

\textbf{Space Aggregation Module.} SAM is proposed to combine luminance prior and chrominance prior based on Equation (10). It takes the outputs of LOS and COS as input to produce the enhanced normal-light image. Specifically, it firstly concatenate $\mathbf{x}^{k}_{lum}$ and $\mathbf{x}^{k}_{chrom}$. Then, a $3\times3$ convolution with $n$ channels is used to combine them. After that, a CAB is added to emphasize the significant features. Finally, we yield the enhanced images in this stage via a $3\times3$ convolution with $3$ channels. SAM is shown in Fig. 5(c) and can be written as below:
\begin{equation} 
\mathbf{x}^{k} = \text{SAM}(\mathbf{x}^{k}_{lum}, \mathbf{x}^{k}_{chrom}).
\end{equation} 


\subsection{Loss Function}
Given $N$ pairs of low/normal-light images $\{ \mathbf{y}_{i},\widetilde{\mathbf{x}}_{i}\},i=[1,2,\cdots,N]$, we can train our proposed DASUNet via the following loss function:
\begin{equation} 
\mathfrak{L} = \frac{1}{N} \sum_{i=1}^{N}  \sum_{j=1}^{k} w^{j} \mathbb{C}(\mathbf{x}_{i}^{j}, \widetilde{\mathbf{x}}_{i}),
\end{equation}
where $k$ is the stage number, $w^{j}$ is the weighting parameter of the $j$-th stage, $\mathbb{C}(\cdot , \cdot)$ indicates the Charbonnier loss, and $\mathbf{x}_{i}^{j}$ denotes the enhanced image of j-th stage of DASUNet. In our training, we set $w^{k}$ = 1 and other $w^{j}$ = 0.1.

%% file: sec/4_results.tex
\section{Experiments and Results}

\subsection{Training Detail and Benchmark}
Our DASUNet is conducted in PyTorch platform with an RTX 3090 GPU. The channel number $n$ for all module (without clear statement) is set to $64$ and the stage number $k$ is initialized as $4$ in our experiments. Adam optimizer with default setting ($\beta_1=0.9$ and $\beta_2=0.99$) is used as the training solver of DASUNet. The initial learning rate is specified as $2\times 10^{-4}$ and gradually decreases to $1\times 10^{-6}$ during $500$ epochs via the cosine annealing strategy. Moreover, we also use the warming up strategy ($3$ epochs) to steady the training of our DASUNet. Training samples are randomly cropped with size of $256\times 256$ and then rotated and flipped for data augmentation. For convenience, images are firstly transformed to YCbCr space for training or inference and are finally re-transformed back to RGB domain for impartial evaluation.

We evaluate the effectiveness of our DASUNet on two popular benchmarks: LOL ~\cite{28} and MIT-Adobe FiveK ~\cite{62}. LOL dataset is comprised of $500$ paired of low/normal-light images, in which $485$ pairs are used for training and the rest is used for test. MIT-Adobe FiveK dataset contains $5000$ low-light images from various real scenes. Like other methods, we use the images retouched by expert-C as reference images. The first $4500$ pairs images are set as the training set and the last $500$ pairs images are selected as test set. Three full-reference evaluators, PSNR, SSIM, and LPIPS ~\cite{63}, are selected as objective metrics. More datasets, including LOL-v2 and some unpaired datasets, and their results are shown in Supplementary Material.

\begin{table*}[!t] \small 
\caption{The comparison results of ours with other state-of-the-art methods on LOL dataset. Blod indicates the best results and underline marks the second-best results.}
\begin{center}
\begin{tabular}{ccc ccccc cc}
\toprule  
Methods & AGCWD  & SRIE   & LIME        & ROPE   & Zero-DCE & RetinexNet & Enlighten-GAN & KinD  & KinD++\\ 
\midrule
PSNR    & 13.05       & 11.86   & 16.76       & 15.02     & 14.86        & 16.77          & 17.48     & 20.38   & 21.80 \\
SSIM     & 0.4038     & 0.4979 & 0.5644     & 0.5092   & 0.5849      & 0.5594         & 0.6578  & 0.8045 & {0.8316}\\
LPIPS   & 0.4816    & 0.3401 & 0.3945      & 0.4713    & 0.3352      & 0.4739        & 0.3223   & 0.1593   & 0.1584    \\ 
\midrule
Methods     & MIRNet & URetinex & Bread   & DCCNet     & LL-Former  & UHDFour &RetinexFormer &LLDiffusion & Ours\\ 
\midrule
PSNR           & \underline{24.14} & 21.33  & 22.96  & 22.98  & 23.65         & 23.10   &25.16&24.65& \textbf{26.60} \\
SSIM       & 0.8302      & 0.7906          & 0.8121 & 0.7909 & 0.8102    & 0.8208  &\underline{0.8434}& 0.8430 & \textbf{0.8552} \\
LPIPS      & 0.1311      & \underline{0.1210} & 0.1597 & 0.1427 & 0.1692    & 0.1466 &0.1314&\textbf{0.0750} & {0.1275} \\
\bottomrule
\end{tabular}
\end{center}
\end{table*}

\begin{table*}[!t] 
\caption{The comparison results of ours with other state-of-the-art methods on MIT-Adobe FiveK dataset. Blod indicates the best results and underline marks the second-best results.}
\begin{center}
\begin{tabular}{ccc ccccc cc}
\toprule   
Methods &SRIE        & LIME     & IAT       &DSN      &DSLR    & SCI       & CSRNet  & MIRNet   & Ours             \\ 
\midrule
PSNR    & 18.44       & 13.28     & 18.08   & 19.58    &20.26    & 20.73    & 20.93      & \underline{23.78}      & \textbf{28.34} \\
SSIM     & 0.7905     & 0.7276   & 0.7888  & 0.8308  &0.8150  & 0.7816  & 0.7875    & \underline{0.8995}    & \textbf{0.9140} \\
LPIPS    & 0.1467     & 0.1573   & 0.1364  & 0.0874 &0.1759   & 0.1066 & 0.1297     & \underline{0.0714}    & \textbf{0.0476} \\
\bottomrule
\end{tabular}
\end{center}
\end{table*}

\subsection{Comparison with State-of-the-Art Methods}

To evaluate the performance of our proposed DASUNet, we compare our results with seventeen classical low-light image enhancement methods which include four traditional models (AGCWD~\cite{66}, SRIE~\cite{15}, LIME~\cite{13}, and ROPE~\cite{67}), two unsupervised methods (Zero-DCE~\cite{27} and EnlightenGAN~\cite{26}), five recent deep black-box methods (MIRNet~\cite{64}, Bread~\cite{35}, LLFormer~\cite{25}, UHDFour~\cite{43}, and LLDiffusion~\cite{83}) and four deep Retinex-based methods (RetinexNet~\cite{28}, KinD~\cite{29}, KinD++~\cite{65}, URetinex~\cite{33}, and RetinexFormer~\cite{84}), on LOL dataset. Quantitative comparison results are summarized in Table 1. It can easily observe that our method outperforms other comparison methods in PSNR and SSIM and achieves the comparative results in LPIPS evaluator. Qualitative results are presented in Fig. 5. As can be seen, RetinexNet cannot achieve expected enhancement images and KinD produces over-smoothing restored images. URetinex shows the color deviation on the ground. LLFormer and RetinexFormer seem to be under-enhanced and noisy. However, the enhanced result by our DASUNet shows more vivid details and more natural scene light, which remove some artifacts and noise that cannot be eliminated by other methods.

Moreover, we conduct experiments on MIT-Adobe FiveK dataset to demonstrate the generalization capability of our proposed DASUNet. We compare our model with eight state-of-the-art methods including SRIE~\cite{15}, LIME~\cite{13}, IAT~\cite{40}, SCI~\cite{45}, DSN~\cite{41}, DSLR~\cite{68}, CSRNet~\cite{44}, and MIRNet~\cite{64} in Table 2. As tabulated in Table 2, our method significantly surpasses the second-best methods, MIRNet, in terms of PSNR, SSIM, and LPIPS. To show the effectiveness of our DASUNet, visual comparisons are manifested in Fig. 6. One can observe that SRIE and DSN connot achieve favorable enhancements. CSRNet overenhances the low-light images. MIRNet encounters ring artifacts in the second row of image. Overall, our DASUNet produces impressive enhanced images. 

\begin{figure*}
  \centering
  \centerline{\includegraphics[width=0.92\textwidth]{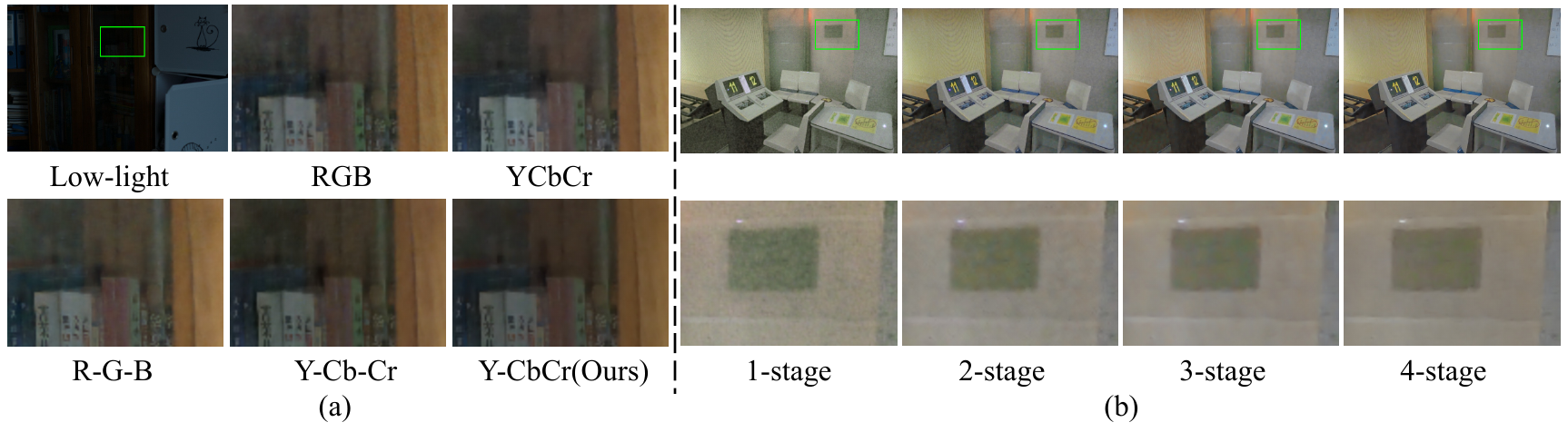}}
  \caption{The visual comparison of ablation study. (a) The visual comparison of different degradation models; (b) The visual comparison of different stages.}
\end{figure*}

\begin{table*}[!t] 
\caption{The comparison results of different degradation models on LOL dataset.}
\begin{center}
\begin{tabular}{c c c  ccc c c c}
				\toprule
				 Methods  & Single model   & Single model  & Double model   &Triple model &Triple model  \\ \midrule 
                    Space    &  RGB&YCbCr&Y-CbCr&R-G-B&Y-Cb-Cr\\
\midrule 
                 PSNR $\uparrow$ 			 & 25.56	&25.62	&\bf26.60	& 24.39     &25.77                \\ 
                 SSIM $\uparrow$ 		     &0.8367 &0.8383	&\bf0.8552	&0.8299    &0.8420                   \\ 
                 LPIPS $\downarrow$ 	     &0.1475 &0.1410	&\bf0.1275	&0.1627    &0.1386                \\ 
                 Time (s)     		                 &\bf0.42	    &\bf0.42	    &0.60	    &0.79        &0.79                \\ 
				\bottomrule
	\end{tabular}
\end{center}
\vspace*{-3mm}
\end{table*}

\begin{figure}
  \centering
  \centerline{\includegraphics[width=0.45\textwidth]{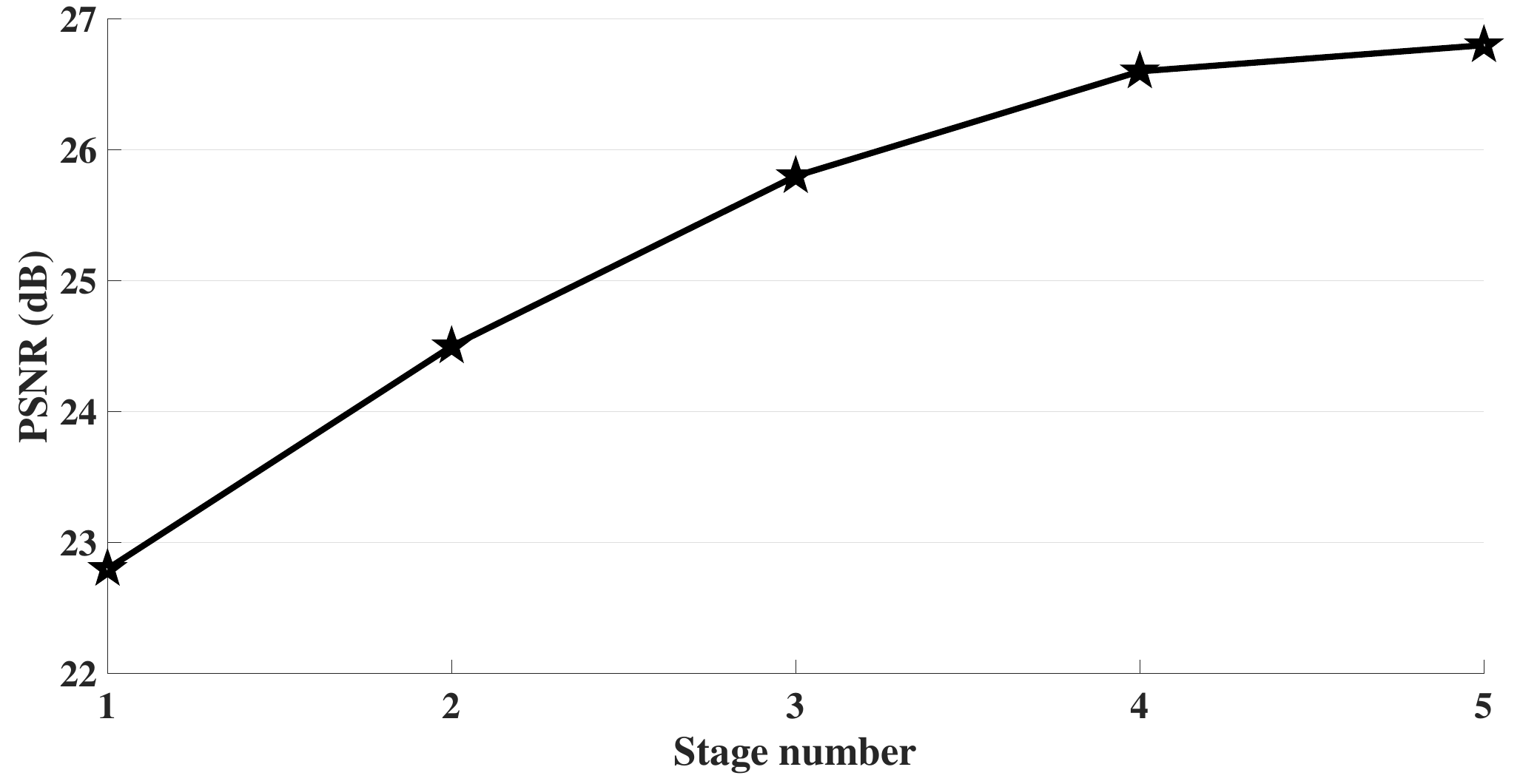}}
  \caption{Ablation study of stage number}
\end{figure}

\subsection{Ablation study}
\textbf{Dual degradation model.} Based on the degradation specificity between luminance and chrominance spaces, we proposed a DDM for low-light image enhancement. To demonstrate its effectiveness, we conduct some comparison experiments on various color spaces and degradation models on LOL dataset, the results of which are presented in Table 3. Single model and triple model denotes one degradation model and three degradation models on corresponding sapces. One can see from Table 3 that the design philosophy behind DDM is effective. Single model cannot consider the the degradation specificity on different spaces and triple model could lose the mutual benefits between homogeneous degradation spaces. Visual comparison of different degradation models are illustrated in Fig. 7(a). Single model introduces some visual artifacts and triple model yields some blurs, while our model produce clearer result.

\textbf{Stage number.} It is well-known that the stage number $k$ is essential for iteration optimization solution. We explore the performance difference of different stage numbers and experimental results are delineated in Fig. 8. When $k$ exceeds $4$, the enhancement performance reaches convergence. Consequently, the stage number in our experiments is set to $4$. Visual comparsion are shown in Fig. 7(b). One can see clearer results with stage advancement.

\subsection{Running Time}
We report the average running time of our method and recent proposed deep learning-based methods on LOL dataset in Fig. 9. Also, we list model parameters of all test methods in Fig. 9. Although LLFormer and MIRNet can achieve noticeable performance, they are heavy in model size. URetinex and Zero-DCE show compact model design, but their performance is poorer than ours. One can see that our method achieves the best performance with an acceptable running time and a relatively economic model size. It embodies a favorable trade-off between the effectiveness and efficiency of our proposed model.

\begin{figure}[!t] 
  \centering
  \centerline{\includegraphics[width=0.48\textwidth]{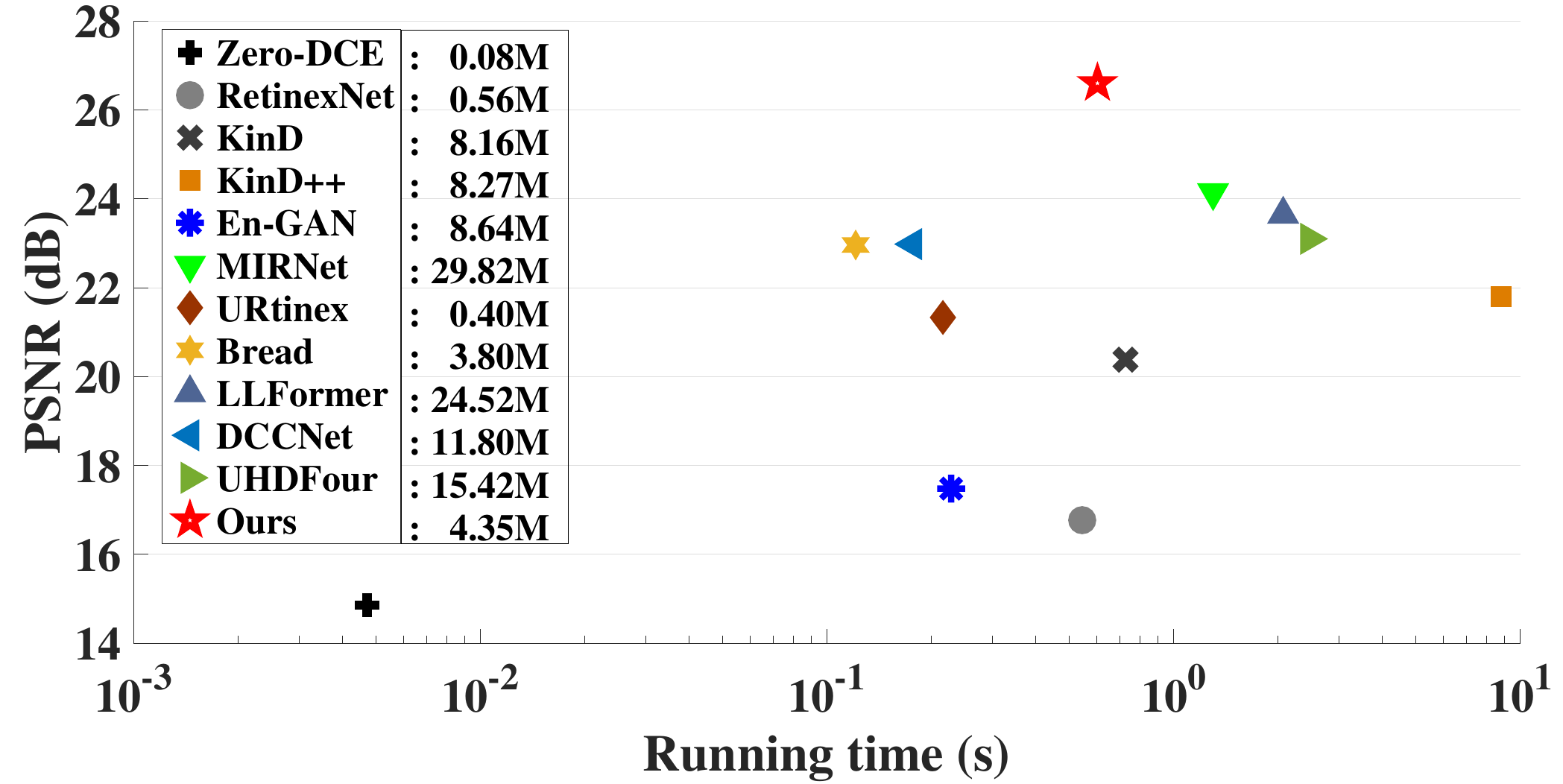}}
  \caption{The comparison on running time and parameters.}
\vspace*{-3mm}
\end{figure}

%% file: sec/5_conclusion.tex
\section{Conclusion}

In this paper, we have proposed a dual degradation-inspired deep unfolding method (DASUNet) for low-light image enhancement. Specifically, we design a dual degradation model (DDM) based on the degradation specificity among luminance and chrominance spaces. An alternative optimization solution is proposed to solve it and is unfolded into specified network modules to construct our DASUNet, which contains a new prior modeling module to capture the long-range prior information and a space aggregation module to combine dual degradation priors. Extensive expeirmental results validate the superiority of our DASUNet for low-light image enhancement. In future, we will explore more low-level vision tasks based on DDM.

%% file: sec/X_suppl.tex
\clearpage
\setcounter{page}{1}
\maketitlesupplementary

\section{Supplementary materials}

\subsection{Dataset detail}
LOL-V2 dataset ~\cite{22} contains 789 pairs of normal/low-light images, in which 689 pairs are used  to train the enhancement model and 100 pairs are used as testset. MEF ~\cite{80} and DICM ~\cite{81} are two unpaired datasets without reference images. They are respectively composed of 17 and 69 images, which usually are used to evaluate the generalization ability of enhancement algorithms. For unpaired datasets, we introduce a no-reference evaluator NIQE ~\cite{82} to assess their performance.

\subsection{Ablation study}

In PMM, we design new luminance adjustment Transformer (LAT) and Wavelet decomposition Transformer (WDT) based on different characteristics of different spaces. To analyze their effects, we perform a set of ablation experiments via breaking them down. As listed in Table 1, the performance will be enhanced by 1.57 dB and 1.72 dB compared to the baseline when adding LAT and WDT. Finally, LAT and WDT jointly achieve the best performance.

\begin{table} 
\caption{The ablation study of luminance adjustment Transformer (LAT) and Wavelet decomposition Transformer (WDT) on LOL dataset.}
\begin{center}
\begin{tabular}{ ccc c c }
				\toprule
				LAT &WDT&PSNR$\uparrow$ &SSIM$\uparrow$  &LPIPS $\downarrow$ \\ 
                \midrule
                &&24.05	&0.8338	&0.1426\\
                \checkmark&&25.62	&0.8415	&0.1387   \\
                &\checkmark&25.77	&0.8437	&0.1354   \\ 	
                \checkmark&\checkmark&\bf26.60	&\bf0.8552	&\bf0.1275   \\
				\bottomrule
	\end{tabular}
\end{center}
\end{table}

Besides, we study the effect of SAM and high-way feature path. As shown in Table 2, the performance will be heavily decreased when deleting SAM and high-way feature path. It demonstrates the importance of SAM and high-way feature path for space prior combination and feature flow.

\begin{table} 
\caption{The ablation study of feature path and SAM on LOL dataset.}
\begin{center}
\begin{tabular}{ ccc c c }
				\toprule
				SAM&feature path&PSNR$\uparrow$ &SSIM$\uparrow$  &LPIPS $\downarrow$ \\ 
                \midrule
                \checkmark&&24.36	&0.8332	&0.1410   \\
                &\checkmark&24.32	&0.8463	&0.1366   \\ 	
                \checkmark&\checkmark&\bf26.60	&\bf0.8552	&\bf0.1275   \\
				\bottomrule
	\end{tabular}
\end{center}
\end{table}

\subsection{Comparisons}

We compare our results with seven methods, including SRIE ~\cite{15}, LIME ~\cite{13}, EnlightenGAN ~\cite{26}, Zero-DCE ~\cite{27}, URetinex ~\cite{33}, SNRANet ~\cite{24}, UHDFour ~\cite{43}, RetinexFormer ~\cite{84}, LLDiffusion ~\cite{83}, and ACCA ~\cite{86}, on LOL-V2 dataset. As shown in Table 3, we outperform other comparison methods in PSNR and SSIM. Visual comparisons are shown in Fig. 1. As shown in Fig. 1, LIME and URetinex over-enhance the low-light images. Zero-DCE produces under-enhanced results. UHDFour encounters the color cast in the first-line image. Our method yields visual-pleased results. It demonstrates the advantages of WDT and LAT.

\begin{table*}[h] \footnotesize
\begin{center}
\begin{tabular}{ccc ccc ccc ccc}

\toprule   
Methods &SRIE        & LIME     & En-GAN       &Zero-DCE      &URetinex    & SNRANet       & UHDFour & RetinexFormer & LLDiffusion & ACCA & Ours             \\ \midrule[.1em]
PSNR    & 14.45       & 15.24     & 18.64   & 18.06    &19.78    & 21.41    & {21.78}    & 22.80 & 23.16 & \underline{23.80}  & \textbf{27.64} \\
SSIM     & 0.5240    & 0.4151   & 0.6767  & 0.5795  &0.8427  & 0.8475  & \underline{0.8542}  & 0.8400 & 0.8420 & 0.8290    & \textbf{0.8659} \\
\bottomrule
\end{tabular}
\end{center}
\caption{The comparison results of ours with other state-of-the-art methods on LOL-V2 dataset. Blod indicates the best results and underline marks the second-best results.}
\label{tab:my-table}
\end{table*}

\begin{figure*}[h]
	\begin{center}
		\begin{tabular}[t]{c} 
			\includegraphics[width=0.98\textwidth]{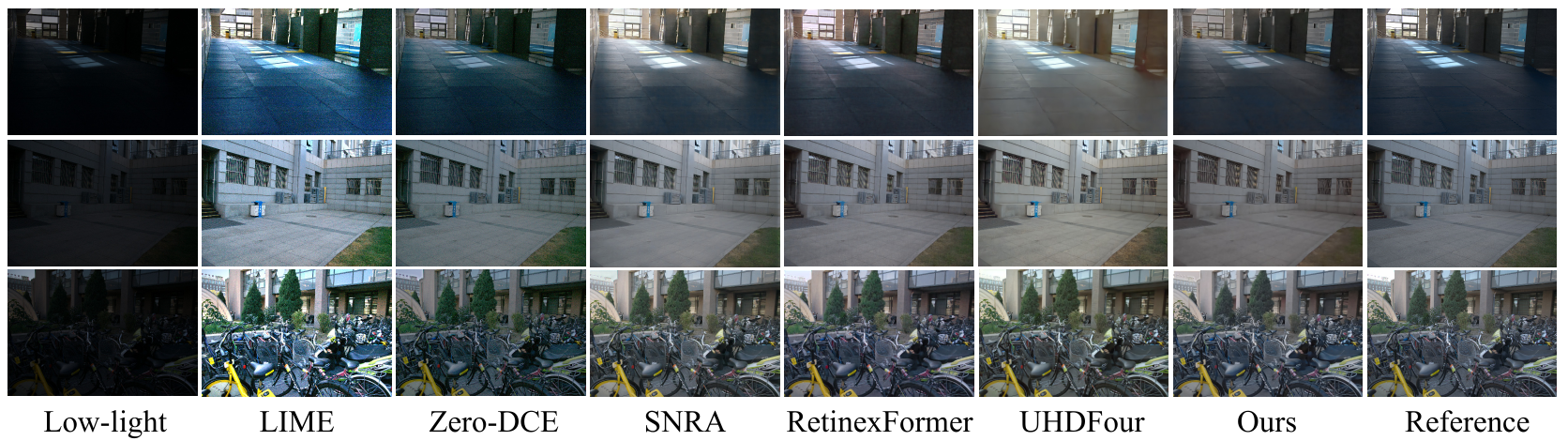}
		\end{tabular}
	\end{center}
	\caption{Visual comparisons on LOL-V2 dataset.}
\end{figure*}

Besides, we compare our results with SRIE ~\cite{15}, Zero-DCE ~\cite{27}, RetinexNet ~\cite{28}, KinD ~\cite{29}, MIRNet ~\cite{64}, Bread ~\cite{35}, and LLFormer ~\cite{25} on MEF and DICM datasets. As shown in Table 4, we obtain the best performance on two datasets in terms of NIQE. Visual comparisons are shown in Fig. 2 and 3. Retinex produces unnatural looks, and KinD achieves under-enhanced results. While MIRNet over-enhances low-light images. LLFormer produces checkboard artifacts.  It can be observed that our method yields impressive enhancement images.

\begin{table*}[!h] \footnotesize
\begin{center}
\begin{tabular}{cccc ccccc }

\toprule   
Datasets &SRIE       & Zero-DCE & RetinexNet     & KinD       &MIRNet      &Bread    & LLFormer      & Ours             \\ \midrule[.1em]
MEF    & 3.2041   &3.3088    & 4.9043     & 3.5598       & 3.1915    & 3.5677      & 3.2847      & \textbf{3.0358} \\
DICM     & 3.3657 &3.0973    & 4.3143     &3.5135  & 3.1533  & 3.4063    & 3.5154    & \textbf{2.9591} \\

\bottomrule
\end{tabular}
\end{center}
\caption{The comparison results of ours with other state-of-the-art methods on MEF and DICM datasets in terms of NIQE.}
\label{tab:my-table}
\end{table*}

\begin{figure*}[!h]
	\begin{center}
		\begin{tabular}[t]{c} 
			\includegraphics[width=0.98\textwidth]{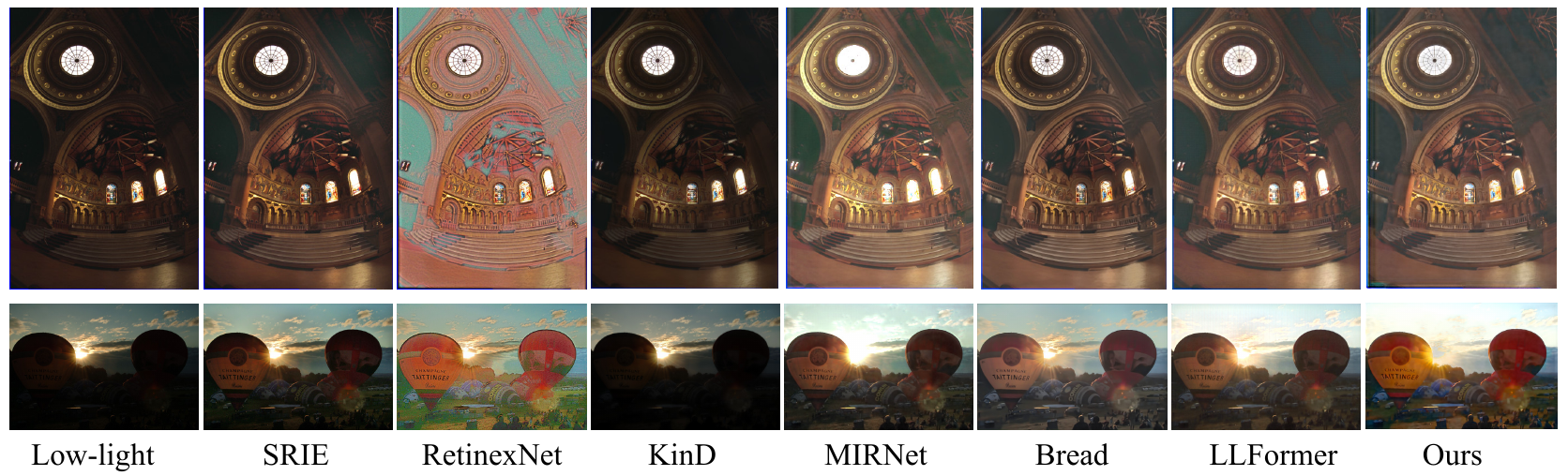}
		\end{tabular}
	\end{center}
	\caption{Visual comparisons on MEF dataset.}
\end{figure*}

\begin{figure*}[!]
	\begin{center}
		\begin{tabular}[t]{c} 
			\includegraphics[width=0.98\textwidth]{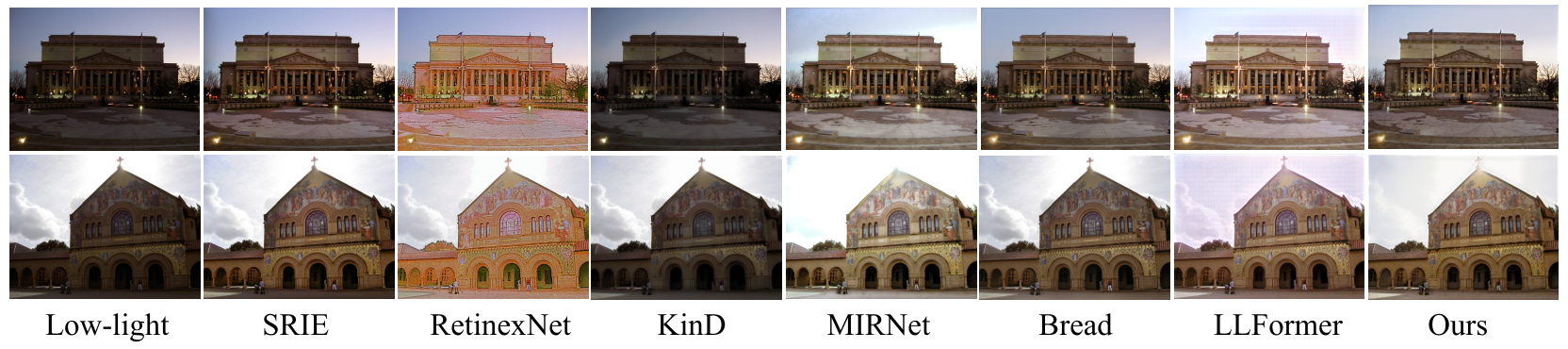}
		\end{tabular}
	\end{center}
	\caption{Visual comparisons on DICM dataset.}
\end{figure*}